\documentclass[letterpaper]{article}
\usepackage{aaai2027}
\nocopyright
\usepackage[hyphens]{url}
\usepackage{graphicx}
\usepackage{natbib}
\usepackage{caption}
\usepackage{amsmath}
\usepackage{amssymb}
\usepackage{booktabs}
\usepackage[table]{xcolor}
\usepackage{tabularx}
\usepackage{array}

\newcommand{\benchmark}{\textsc{CircuitReason-1k}}
\newcommand{\cmark}{\(\checkmark\)}
\newcommand{\xmark}{\(\times\)}
\newcolumntype{Y}{>{\raggedright\arraybackslash}X}

\title{CircuitReason-1k: Benchmarking Long-Horizon Visual-to-Symbolic Reasoning in Electrical Circuits}
\author{
    Xinqi Yang\textsuperscript{\rm 1}\equalcontrib,
    Kang An\textsuperscript{\rm 1}\equalcontrib\thanks{Project leader},
    Tengyue Wang\textsuperscript{\rm 4},
    Zhongyu Yang\textsuperscript{\rm 2},
    Chenxu Du\textsuperscript{\rm 5},\\
    Yuanchi Zhu\textsuperscript{\rm 6,\rm 7},
    Hebao Zhu\textsuperscript{\rm 8},
    Ziliang Wang\textsuperscript{\rm 3},
    Faqiang Qian\textsuperscript{\rm 3},
    Yunli Yang\textsuperscript{\rm 9},
    Qibing Ren\textsuperscript{\rm 1}\corresponding
}
\affiliations{
    \textsuperscript{\rm 1}Shanghai Jiao Tong University,
    \textsuperscript{\rm 2}ModelBest,
    \textsuperscript{\rm 3}SenseTime,
    \textsuperscript{\rm 4}South China University of Technology,\\
    \textsuperscript{\rm 5}Southwest Jiaotong University,
    \textsuperscript{\rm 6}ShanghaiTech University,
    \textsuperscript{\rm 7}Institute of Automation, Chinese Academy of Sciences,\\
    \textsuperscript{\rm 8}Chongqing University,
    \textsuperscript{\rm 9}Institute for Advanced Algorithms Research, Shanghai\\
    \{an\_kang, renqibing\}@sjtu.edu.cn
}

\begin{document}

\maketitle

\begin{abstract}
Electrical circuit analysis requires more than recognizing components in an
image. A solver must ground symbols and labels, recover latent topology,
select a physical model, formulate coupled equations, propagate intermediate
quantities, and preserve units, signs, directions, and phase conventions.
We introduce \benchmark, a benchmark of 1,000 authentic textbook problems for
evaluating this complete long-horizon visual-to-symbolic reasoning process.
Each problem pairs one or more circuit diagrams with a self-contained question,
a typed or semantically specified answer, and a reference worked solution.
An evidence-first construction pipeline aligns questions, figures, and
solutions, while a reasoning-oriented taxonomy organizes problems by circuit
type and dependency depth. Evaluation combines conservative typed scoring with
identity-blinded multi-model semantic consensus, retaining every problem in
the denominator. Across three commercial chatbot systems and six open-source
multimodal large language models,
the highest-scoring system reaches 84.8\% accuracy. However, performance consistently
deteriorates on long-horizon problems, and qualitative analysis exposes
persistent failures in topology-to-target binding, physical conventions, and
late-stage output propagation. \benchmark{} provides a focused testbed for
measuring whether multimodal models can transform technical visual evidence
into sustained, physically valid symbolic reasoning. Code are available at
\url{https://github.com/CircuitReason/CircuitReason1K}.
\end{abstract}

\section{Introduction}

Electronic and semiconductor systems underpin enormous global markets in
computing, communications, energy, transportation, and industrial control.
Circuit-diagram understanding, analysis, verification, and troubleshooting
recur across development, consuming substantial engineering time and R\&D cost
through repeated schematic interpretation, topology recovery, physical
modeling, equation derivation, and result verification. Reliable multimodal
circuit reasoning could relieve this costly bottleneck in AI-assisted circuit
engineering, design verification, debugging, and education.

\noindent\begin{minipage}{\columnwidth}
    \centering
    \includegraphics[width=0.96\columnwidth]{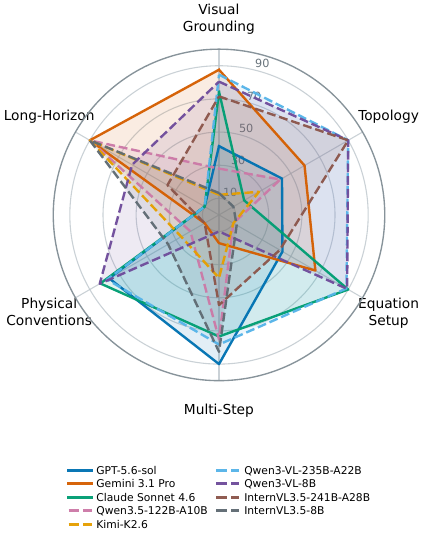}
    \captionof{figure}{Within-model diagnostic profiles. Radii are normalized
    within each model and show relative, not absolute, strengths; cross-model
    comparisons require Table~\ref{tab:main-results}. Scores are z-normalized
    by dimension across models, then min--max scaled to $[10,90]$ within each
    model. Solid and dashed outlines denote commercial and open-source systems.}
    \label{fig:diagnostic-radar}
\end{minipage}
\par\vspace{0.5\baselineskip}

Recent multimodal benchmarks, including ScienceQA~\cite{lu2022scienceqa},
MathVista~\cite{lu2024mathvista}, and MMMU~\cite{yue2024mmmu}, have
advanced visual and scientific reasoning evaluation, yet provide limited
coverage of electrical circuits. Their circuit questions are often sparse or
solvable through localized cues and short calculations. Circuit-oriented
datasets, meanwhile, primarily emphasize component recognition, diagram
parsing, or synthetic generation. These settings do not systematically test
the complete perception--topology--equation--solution chain found in authentic
circuit analysis. It therefore remains unclear whether multimodal large
language models (MLLMs) can sustain physically valid reasoning over complex
diagrams rather than recognize familiar patterns.

In engineering practice, a circuit diagram is not merely a collection of
recognizable objects, but a relational system whose meaning depends on
connectivity, orientation, polarity, reference direction, and component
values. Engineers must interpret symbols and labels, distinguish crossings
from junctions, reconstruct latent topology, select physical laws, formulate
coupled equations, propagate intermediate quantities, and verify a
dimensionally and conventionally valid result.

We use \emph{long-horizon} to denote a sustained chain of interdependent
visual, structural, mathematical, and physical reasoning steps. Because each
stage conditions the next, one local grounding or convention error can
invalidate an otherwise correct derivation.
For example, finding one branch current may require recovering the global node
graph, deriving an equivalent impedance, solving coupled voltages, and only
then applying the queried reference direction. A benchmark intended to support
real circuit engineering should therefore evaluate this complete chain rather
than isolated recognition or short-answer calculation.

\begin{figure*}[t]
    \centering
    \includegraphics[width=0.95\textwidth]{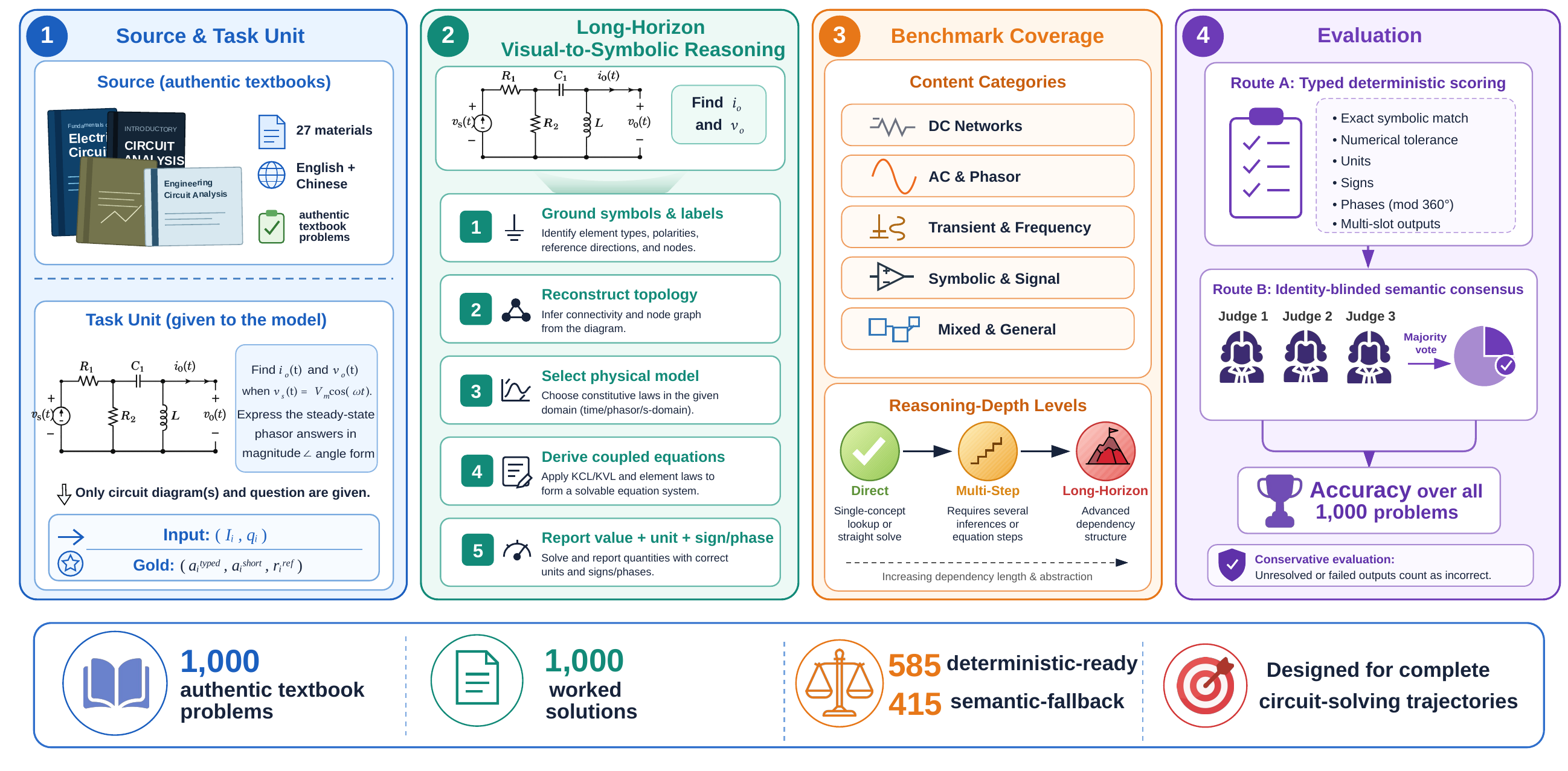}
    \caption{Overview of \benchmark. A model receives an authentic circuit
    diagram and a self-contained question, then must sustain a long-horizon
    visual-to-symbolic chain from grounding and topology reconstruction to
    physical modeling, coupled equations, and a convention-complete answer.}
    \label{fig:overview}
\end{figure*}

To address this gap, we introduce \benchmark, a dedicated benchmark for
evaluating this complete reasoning process. It contains 1,000 carefully
curated problems from authentic university-level textbooks and problem
collections. Each instance pairs one or more circuit diagrams with a
self-contained question, a typed or semantically specified answer, and a
detailed worked solution. The problems span DC networks, nodal and mesh
analysis, network theorems, operational amplifiers, transient response,
sinusoidal steady state, phasors, complex power, and frequency-domain
reasoning. They range from visually sensitive calculations to long
derivations with coupled quantities and transformations among graphical,
algebraic, and physical representations. A reasoning-oriented taxonomy
characterizes circuit type, visual complexity, topology reconstruction,
dependent reasoning hops, and mathematical burden, supporting both aggregate
evaluation and fine-grained failure analysis. An evidence-first construction
pipeline binds each question to its source diagram and reference derivation
before role-separated verification.

Reliable evaluation of circuit answers is itself nontrivial. Semantically
equivalent outputs can differ in units, scaling, notation, sign conventions,
or algebraic form. We therefore combine conservative typed deterministic
scoring with identity-blinded, multi-model semantic consensus for complex
cases, while retaining missing responses and generation failures in the
denominator. Across three commercial chatbot systems and six open-source
MLLMs, the strongest
system reaches 84.8\%, but accuracy consistently deteriorates on long-horizon
problems, with persistent errors in topology-to-target binding, physical
conventions, and late-stage answer propagation. Table~\ref{tab:main-results}
reports absolute performance, while Figure~\ref{fig:diagnostic-radar} exposes
complementary model-specific strengths and weaknesses.

Our main contributions are summarized as follows:
\begin{itemize}
    \item We introduce \benchmark, a benchmark of 1,000 authentic textbook
    problems for evaluating long-horizon visual-to-symbolic reasoning over
    electrical circuit diagrams.

    \item We provide multi-granular answers, worked solutions, and a
    reasoning-oriented taxonomy that makes dependency depth and failure modes
    observable.

    \item We establish a hybrid evaluation protocol combining typed scoring
    with independent multi-model consensus, and benchmark nine leading MLLMs
    under standardized reasoning configurations.
\end{itemize}

\section{Related Work}

\subsection{Multimodal Scientific and Mathematical Reasoning}

Scientific diagrams have long tested reasoning beyond natural-image
recognition. CLEVR and NS-VQA study compositional visual reasoning
~\cite{johnson2017clevr,yi2018nsvqa}; AI2D and IconQA target diagram
understanding~\cite{kembhavi2016ai2d,lu2021iconqa}; chart benchmarks emphasize
structured numerical evidence~\cite{kafle2018dvqa,methani2020plotqa,
masry2022chartqa}; and GeoQA and Inter-GPS couple geometry diagrams to
symbolic procedures~\cite{chen2021geoqa,lu2021intergps}.

ScienceQA~\cite{lu2022scienceqa}, MathVista~\cite{lu2024mathvista},
MMMU~\cite{yue2024mmmu}, MathVerse~\cite{zhang2024mathverse}, and
OlympiadBench~\cite{he2024olympiadbench} broaden multimodal scientific and
mathematical evaluation. Perception-focused diagnostics such as
BLINK~\cite{fu2024blink}, HallusionBench~\cite{guan2024hallusionbench}, and
MMStar~\cite{chen2024mmstar} further motivate explicit visual-dependence checks
and fine-grained failure analysis.

These benchmarks establish the importance of combining perception with
mathematical and scientific reasoning, but their broad coverage provides only
limited resolution on electrical circuits. Circuit analysis introduces a
particularly strict dependency structure: models must recover connectivity and
reference conventions before selecting physical laws, constructing equations,
and propagating intermediate quantities. This differs from solving a problem
after its mathematical representation has already been supplied.
\benchmark{} focuses on this complete visual-to-symbolic chain and supports
analysis by circuit type, answer form, and reasoning horizon.

\textbf{Open-ended evaluation.}
MM-Vet evaluates heterogeneous answers with an LLM judge~\cite{yu2024mmvet};
broader studies document both scalability and systematic
biases~\cite{zheng2023llmjudge,liu2023geval}. This motivates typed rules where
possible, identity blinding, independent votes, and strict-majority consensus.

\subsection{Circuit Diagram Understanding and Circuit-Centric Benchmarks}

Prior work on circuit images has primarily emphasized recognition and
digitization. Bohara and Krishnamoorthy~\cite{bohara2024powerconverter} combine
component detection, wire tracing, and connectivity extraction to convert
hand-drawn power-converter schematics into simulation-ready netlists. Such
systems address the important perception-to-structure stage, but do not
evaluate whether a general-purpose multimodal model can derive and execute a
complete circuit-analysis solution. CircuitVQA~\cite{mehta2024circuitvqa}
moves closer to language-based evaluation with more than 115K questions over
schematic and hand-drawn circuit images. Its automatically instantiated
questions predominantly test component counting, values, spatial relations,
positions, and junctions, making it well suited to measuring circuit
perception but less focused on long, dependent solution trajectories.

The most closely related benchmark is
CircuitSense~\cite{akbari2026circuitsense}, which evaluates perception,
analysis, and design through a combination of curated questions and a
hierarchical synthetic generation pipeline with symbolic equations as ground
truth. CircuitSense demonstrates a pronounced gap between component
recognition and equation derivation, establishing visual-to-mathematical
modeling as a central challenge for engineering MLLMs. CircuitFormer
~\cite{islam2026circuitformer} studies a complementary direction: generating
analog circuit topologies and netlists from natural-language specifications,
with CircuitBench-100 used to evaluate syntactic validity and functional
design.

\begin{table*}[t]
\centering
\small
\setlength{\tabcolsep}{3.6pt}
\begin{tabular*}{\textwidth}{@{\extracolsep{\fill}}lccccccc@{}}
\toprule
\textbf{Benchmark} &
\shortstack{\textbf{Circuit}\\\textbf{analysis}} &
\shortstack{\textbf{Authentic}\\\textbf{circuit text}} &
\shortstack{\textbf{Visual}\\\(\rightarrow\)\textbf{topology}} &
\shortstack{\textbf{Topology}\\\(\rightarrow\)\textbf{equations}} &
\shortstack{\textbf{Long-horizon}\\\textbf{solution}} &
\shortstack{\textbf{Open}\\\textbf{answer}} &
\shortstack{\textbf{Worked}\\\textbf{solution}} \\
\midrule
AI2D & \xmark & \xmark & \xmark & \xmark & \xmark & \xmark & \xmark \\
ScienceQA & \xmark & \xmark & \xmark & \xmark & \xmark & \xmark & \xmark \\
MathVista & \xmark & \xmark & \xmark & \xmark & \xmark & \cmark & \xmark \\
MMMU & \xmark & \xmark & \xmark & \xmark & \xmark & \xmark & \xmark \\
MathVerse & \xmark & \xmark & \xmark & \xmark & \xmark & \cmark & \xmark \\
CircuitVQA & \xmark & \xmark & \cmark & \xmark & \xmark & \xmark & \xmark \\
CircuitSense & \cmark & \xmark & \cmark & \cmark & \xmark & \cmark & \xmark \\
CircuitFormer & \xmark & \xmark & \xmark & \xmark & \xmark & \xmark & \xmark \\
\midrule
\textbf{\benchmark} & \cmark & \cmark & \cmark & \cmark & \cmark & \cmark & \cmark \\
\bottomrule
\end{tabular*}
\caption{Capability coverage of representative multimodal and circuit-centric
benchmarks. ``Long-horizon solution'' denotes evaluation of a dependent chain
that proceeds from diagram evidence through topology and equations to a final
physical answer; ``worked solution'' denotes a complete reference derivation
rather than a short rationale. A check denotes systematic benchmark-level
coverage as a primary design objective; a cross
means that the capability is not systematically evaluated, even if occasional
examples occur. \benchmark{} is the only benchmark in this comparison that
unifies all seven properties.}
\label{tab:related-comparison}
\end{table*}

\benchmark{} differs in its benchmark unit and evaluation objective. It centers
1,000 authentic textbook-style analysis problems whose answers may require
topology recovery, method selection, several coupled equations, and multistage
numerical or symbolic derivations. Detailed worked solutions expose the
dependencies between intermediate quantities, while typed answer schemas and
multi-model judging support reliable evaluation of open-ended answers with
units, tolerances, alternative forms, and semantic equivalence.
\benchmark{} is therefore complementary to perception-oriented CircuitVQA,
synthetic equation-centered CircuitSense, and specification-to-design
CircuitFormer: it targets the end-to-end, long-horizon process of solving
visually grounded circuit-analysis problems.

\section{\benchmark{} Construction}
\label{sec:construction}

\begin{figure*}[t]
    \centering
    \includegraphics[width=0.95\textwidth]{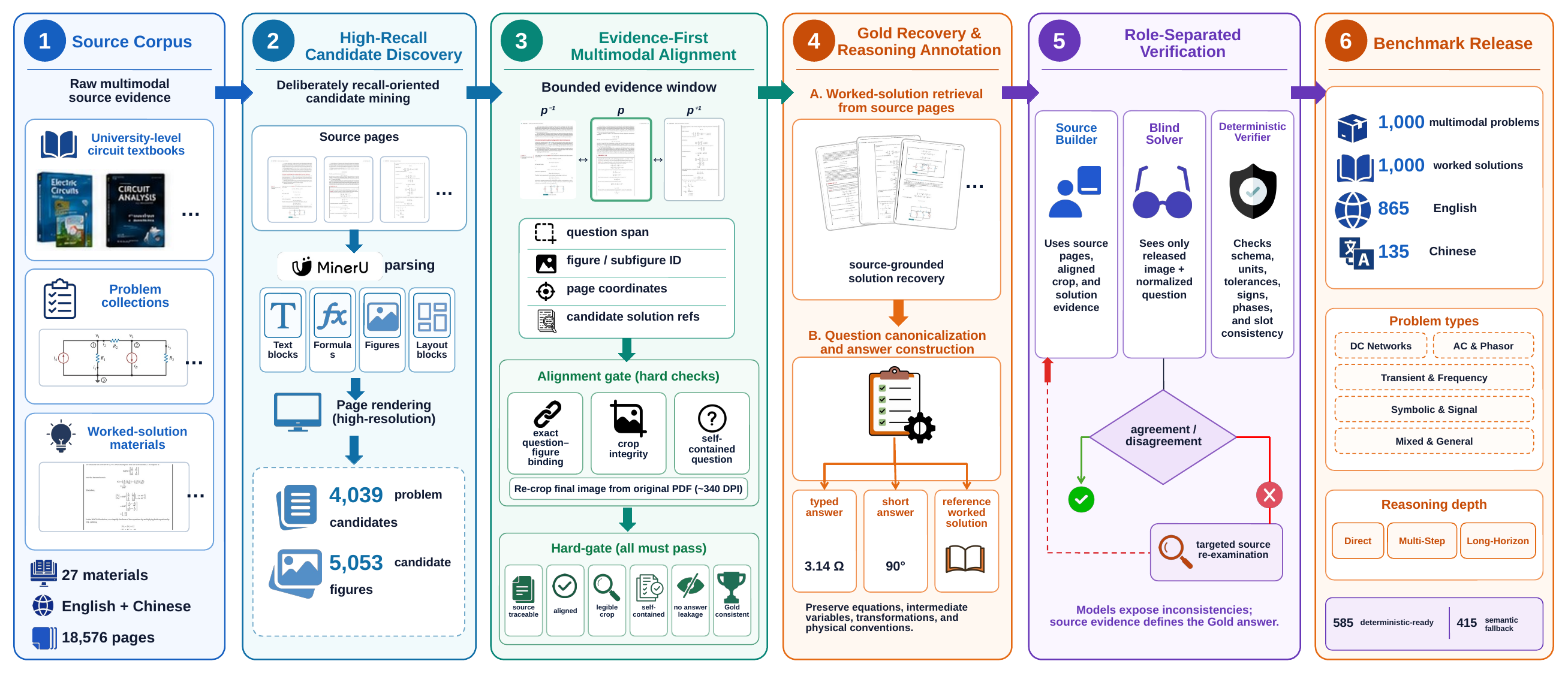}
    \caption{The evidence-first construction pipeline. High-recall document
    mining is followed by exact question--figure--solution alignment,
    multi-granular answer construction, and role-separated checks before the
    final 1,000-problem release.}
    \label{fig:construction-pipeline}
\end{figure*}

\benchmark{} evaluates complete circuit-solving trajectories rather than
isolated recognition or formula retrieval. Each item must consistently link
its diagram, question, reference solution, and answer while remaining
self-contained, visually grounded, and assessable. Our \emph{evidence-first}
pipeline combines high-recall discovery, multimodal alignment,
reasoning-oriented annotation, and role-separated verification
(Figure~\ref{fig:construction-pipeline}).

\subsection{Design Principles and Task Unit}
\label{subsec:construction-principles}

We formulate each instance as
\begin{equation}
    x_i = (I_i, q_i), \qquad
    y_i = \left(a_i^{\mathrm{typed}},
                 a_i^{\mathrm{short}},
                 r_i^{\mathrm{ref}}\right),
    \label{eq:task-record}
\end{equation}
where $I_i$ denotes one or more circuit images and $q_i$ is a self-contained
question. The supervision contains a typed or semantically specified answer
$a_i^{\mathrm{typed}}$, a concise human-readable answer
$a_i^{\mathrm{short}}$, and a reference worked solution
$r_i^{\mathrm{ref}}$. Only $(I_i,q_i)$ is provided to the evaluated model; the
answers and reference solution are withheld.

Three principles guide construction. First, \textbf{authenticity}: problems
should retain the compositional structure of university-level circuit
analysis, including realistic notation, topology, and physical conventions.
Second, \textbf{evidence integrity}: every released question, diagram, and
answer must be traceable to a mutually consistent evidence chain. Third,
\textbf{reasoning observability}: the annotation should preserve intermediate
equations and dependencies needed to analyze a long solution, while retaining
a final answer that can be evaluated reliably. Accordingly,
``long-horizon'' is not defined by response length. It refers to a sequence of
dependent transitions from visual entities to topology, from topology to a
circuit model, and from that model to equations and physically valid outputs.

\subsection{Source Corpus and High-Recall Discovery}
\label{subsec:source-discovery}

We begin with 27 university-level circuit-analysis textbooks, problem
collections, and accompanying solution materials in English and Chinese. The
corpus spans resistive networks, Kirchhoff's laws, nodal and mesh analysis,
network theorems, operational amplifiers, transient response, sinusoidal
steady state, phasors, complex power, three-phase systems, frequency response,
coupled circuits, and two-port networks.

MinerU~\cite{wang2024mineru} recovers text, formula, figure, and layout blocks
with reading order, while high-resolution page renders preserve the original
visual evidence. Candidate mining combines document structure, problem
identifiers, figure references, equation-rich spans, and solution sections.
This stage favors recall: extracted blocks locate candidates but do not
establish question--image or question--solution correspondence.

The initial pass processes 18,576 pages and produces 4,039 problem candidates
and 5,053 candidate figures. This broad search space is subsequently reduced
by the evidence gates described below.

\subsection{Evidence-First Multimodal Alignment}
\label{subsec:evidence-alignment}

\paragraph{Local evidence packets.}
Questions and figures may be separated across pages or surrounded by similar
diagrams. For a question beginning on page $p$, we construct the bounded window
\begin{equation}
    W_p = \{p-1,p,p+1\}.
    \label{eq:evidence-window}
\end{equation}
and extend it once to $p\pm2$ only for an unresolved explicit figure reference.
The packet records the verbatim problem, figure and subfigure identifiers,
coordinates, nearby diagrams, and candidate solution references. This covers
cross-page layouts without permitting unconstrained attachment elsewhere.

\paragraph{Exact question--figure binding.}
Alignment resolves the source problem, referenced figure, and requested
subfigure. Explicit references are hard constraints rather than invitations to
choose the nearest circuit-like image. The final crop is regenerated from the
original PDF at approximately 340\,DPI to preserve values, polarity marks,
phase labels, arrowheads, and junction dots.

We express the hard evidence gate for candidate $c$ as
\begin{equation}
    G_{\mathrm{hard}}(c) =
    G_{\mathrm{src}}
    \land G_{\mathrm{align}}
    \land G_{\mathrm{crop}}
    \land G_{\mathrm{self}}
    \land G_{\mathrm{safe}}
    \land G_{\mathrm{gold}},
    \label{eq:hard-gate}
\end{equation}
where the terms require traceable source evidence, exact question--figure
identity, a complete and legible crop, a self-contained question, absence of
answer leakage, and consistency between the source solution and annotated
answer. Candidates failing a hard condition are rejected or returned for
additional evidence recovery.

\paragraph{Counterfactual visual grounding.}
A positive compatibility judgment alone can be unreliable because many circuit
figures appear superficially similar. We therefore use two counterfactual
diagnostics. In the \emph{image-shuffle test}, an independent verifier receives
the question and a randomly ordered pair consisting of the aligned diagram
$I_i^{+}$ and a domain-matched negative $I_i^{-}$. It must identify
$I_i^{+}$ and verify the requested subfigure, crop completeness, and absence
of leaked answers. In the \emph{text-only test}, the same question is presented
without its diagram to determine whether the topology and parameters have
already been exposed by the text. For a visually grounded item, the desired
relation is
\begin{equation}
    \begin{aligned}
        \operatorname{Match}(q_i,I_i^{+})
        &> \operatorname{Match}(q_i,I_i^{-}),\\
        \operatorname{Answerable}(q_i) &= 0.
    \end{aligned}
    \label{eq:visual-gate}
\end{equation}
These tests distinguish problem-specific visual evidence from the mere
presence of a circuit image and complement source-based verification.

\subsection{Solution Recovery and Reasoning Annotation}
\label{subsec:solution-annotation}

\paragraph{Source-grounded solution recovery.}
Starting from the exact problem identifier, we retrieve solution evidence in
contiguous two-page blocks, up to six pages. Each step checks problem identity,
coverage of requested subparts, completion of a final answer, and the boundary
to the next problem. A final-answer record is not treated as a worked rationale
without further reconstruction and verification.

Source validity and model solvability are separated: blind-solver failure does
not remove an otherwise consistent difficult item. It instead triggers another
independent solution and targeted source inspection, avoiding selection for
problems already solvable by the construction model.

\paragraph{Question canonicalization.}
The source question is transformed into a self-contained prompt by resolving
references such as ``the preceding example,'' making the requested quantity
and reference direction explicit, and separating independently scoreable
subquestions when necessary. Canonicalization is semantics preserving: it may
standardize notation, units, and formatting, but may not change connectivity,
component values, source polarity or phase, initial conditions, or the target
answer.

\paragraph{Multi-granular supervision.}
Each accepted item receives complementary answer views. The \emph{typed
answer} decomposes the target into one or more slots with quantity names,
values or expressions, units, tolerances, and applicable polarity, direction,
or phase references. The \emph{short answer} provides a concise open-ended
target, while the \emph{reference worked solution} records the equations,
intermediate variables, transformations, and physical conventions that connect
the diagram to the final result. For a numerical slot, the canonical
acceptance region is
\begin{equation}
    |\hat{y}-y|
    \leq
    \max\!\left(\tau_{\mathrm{abs}},
                \tau_{\mathrm{rel}}|y|\right).
    \label{eq:numeric-tolerance}
\end{equation}
The reference solution is an auditable problem solution and is never exposed
to the model during benchmark inference.

\subsection{Role-Separated Verification}
\label{subsec:role-verification}

To reduce correlated construction errors, \benchmark{} separates access to
source evidence from independent problem solving. The validation loop contains
three functional roles:
\begin{itemize}
    \item \textbf{Source Builder.} This role receives the source pages,
    aligned crop, and solution evidence. It constructs the canonical question,
    answer representations, and reference solution.

    \item \textbf{Blind Solver.} This role receives only the final diagram and
    normalized question. It cannot access the source solution or Gold answer,
    and independently reconstructs the topology, equations, and final result.

    \item \textbf{Deterministic Verifier.} This role enforces JSON-schema
    validity, image decodability, field completeness, numerical tolerances,
    unit and SI-prefix normalization, slot consistency, supported symbolic
    equivalence, and phase and reference conventions.
\end{itemize}

Agreement supplies positive evidence; disagreement triggers re-examination of
the diagram, equations, conventions, and source pages rather than majority
voting. Textbook and checkable physical evidence define the reference answer;
model roles expose inconsistencies but do not create truth by consensus. A
review interface joins the crop, source pages, prompt, answer views, provenance,
and warnings into an auditable chain.

\subsection{Dataset Composition and Reasoning Depth}
\label{subsec:dataset-composition}

\benchmark{} contains exactly 1,000 multimodal problems (865 English and 135
Chinese), each linked to a worked solution, concise answer, and typed or
semantically specified Gold representation. Answer forms include structured
fields, source-verbatim text, expressions, multiple quantities, scalars, and
short text, reflecting the varied outputs of circuit analysis.

We organize content into five problem types. \emph{DC Networks} covers static
resistive and controlled-source analysis; \emph{AC \& Phasor} covers
sinusoidal steady state and complex power; \emph{Transient \& Frequency}
combines time-domain dynamics with Laplace and frequency-domain analysis;
\emph{Symbolic \& Signal} emphasizes expression derivation and waveform
reasoning; and \emph{Mixed \& General} collects problems whose reasoning spans
several regimes or is not tied to a single operating regime.

Reasoning depth is defined by dependency structure. \emph{Direct} problems
require a short, locally grounded derivation. \emph{Multi-Step} problems
require a routine sequence of topology recovery, equation setup, and
calculation. \emph{Long-Horizon} problems contain multistage or advanced
dependencies: multiple coupled quantities, transformations between
representations, state propagation, or convention-sensitive outputs whose
correctness depends on earlier intermediate results. This definition captures
the trajectory a solver must maintain rather than the number of tokens it
produces.

\begin{table}[t]
\centering
\small
\setlength{\tabcolsep}{5pt}
\begin{tabular}{lr@{\hspace{12pt}}lr}
\toprule
\multicolumn{2}{c}{\textbf{Problem type}} &
\multicolumn{2}{c}{\textbf{Reasoning depth}} \\
\cmidrule(lr){1-2}\cmidrule(lr){3-4}
DC Networks & 206 & Direct & 239 \\
AC \& Phasor & 308 & Multi-Step & 429 \\
Transient \& Frequency & 173 & Long-Horizon & 332 \\
Symbolic \& Signal & 55 & & \\
Mixed \& General & 258 & & \\
\midrule
\multicolumn{4}{l}{\textbf{Answer representation}} \\
Structured & 569 & Source-verbatim & 281 \\
Expression & 66 & Multi-numeric & 62 \\
Numeric & 18 & Text & 4 \\
\midrule
English & 865 & Chinese & 135 \\
\textbf{Problems} & \textbf{1,000} &
\textbf{Worked solutions} & \textbf{1,000} \\
\bottomrule
\end{tabular}
\caption{Composition of \benchmark{} by problem type, reasoning depth, answer
representation, and language.}
\label{tab:dataset-composition}
\end{table}

\section{Evaluation Protocol}
\label{sec:evaluation}

\subsection{Typed Deterministic Scoring}

The first route applies conservative rules to answer structures for which
equivalence is explicitly encoded. These rules support normalized exact
matching, scalar and multi-slot numerical comparison, absolute and relative
tolerances, SI units and prefixes, and required signs, phases, or reference
directions. Problem-level exactness requires every requested output. Unsupported
symbolic equivalence, ambiguous target mapping, or incomplete prose is deferred
rather than guessed. The release contains 585 deterministic-ready problems.

\subsection{Reference-Assisted Multi-Model Consensus}

The remaining 415 semantically rich problems use an identity-blinded rubric:
judges see the Gold answer and solution but not model identity, and compare
equivalence, completeness, units, signs, directions, phases, and constraints.
At least three valid votes and a unique strict majority are required;
malformed, failed, uncertain, or unresolved cases cannot create a positive
decision and are counted as incorrect.

Judge outputs pass a strict parser. Syntax-only recovery repairs illegal JSON
backslashes, quotes, and control characters only if valid JSON results; it
never rewrites equations, values, or natural-language content.

\subsection{Metric}

Let $e_{im}\in\{0,1\}$ indicate that model $m$ is exactly correct on problem
$i$. Missing, failed, and unresolved outputs receive $e_{im}=0$. With
$N=1000$, the primary metric is
\begin{equation}
    \operatorname{Accuracy}_m =
    \frac{1}{N}\sum_{i=1}^{N} e_{im}.
    \label{eq:accuracy}
\end{equation}
This fixed-denominator definition measures end-to-end model behavior: a model
must both produce an answer and make every requested quantity correct. The same
definition is used for all problem-type and reasoning-depth subsets.

\section{Experiments}
\label{sec:experiments}

\subsection{Models and Inference Protocol}

We evaluate nine representative multimodal models. We use the endpoint
identifiers recorded in the frozen evaluation artifacts. The commercial
chatbot systems are GPT-5.6-sol, Gemini 3.1 Pro Preview (abbreviated as Gemini
3.1 Pro in figures and tables), and Claude Sonnet 4.6. The open-source MLLMs are
Qwen3.5-122B-A10B, Kimi-K2.6,
Qwen3-VL-235B-A22B, Qwen3-VL-8B, InternVL3.5-241B-A28B, and
InternVL3.5-8B. Open models use their native thinking or deep-thinking mode.
This selection covers frontier sparse models, large vision-language models,
and two smaller 8B baselines.

All systems receive the same image and question without the Gold answer or
solution. No request-level output-token cap is imposed: local generation uses
the remaining context (up to 131,072 tokens where supported), and commercial
chatbot endpoints use provider-managed maxima. We retain native templates and
recommended decoding, version all responses and scoring records, and use the
full subset denominator for every accuracy.

\begin{table*}[t]
\centering
\scriptsize
\setlength{\tabcolsep}{3.1pt}
\renewcommand{\arraystretch}{0.94}
\resizebox{\textwidth}{!}{%
\begin{tabular}{lrrrrrrrrr}
\toprule
& \multicolumn{5}{c}{\textbf{Problem Type}} &
\multicolumn{3}{c}{\textbf{Reasoning Depth}} & \\
\cmidrule(lr){2-6}\cmidrule(lr){7-9}
\textbf{Model} &
\textbf{DC} & \textbf{AC} & \textbf{T/F} & \textbf{Sym.} &
\textbf{Mix} & \textbf{Direct} & \textbf{Multi} &
\textbf{Long} & \textbf{Overall} \\
\midrule
\rowcolor{black!8}
\multicolumn{10}{c}{\textbf{Commercial Chatbot Systems}} \\
GPT-5.6-sol &
\textbf{86.89} & 83.44 & 83.82 & \textbf{90.91} & 84.11 &
\textbf{85.36} & 88.11 & 80.12 & \textbf{84.80} \\
Gemini 3.1 Pro &
85.92 & 82.47 & 80.92 & 81.82 & 80.23 &
75.31 & 87.65 & 80.42 & 82.30 \\
Claude Sonnet 4.6 &
66.99 & 63.64 & 67.05 & 70.91 & 60.08 &
62.76 & 69.46 & 59.04 & 64.40 \\
\midrule
\rowcolor{black!8}
\multicolumn{10}{c}{\textbf{Open-Source MLLMs}} \\
Qwen3.5-122B-A10B &
\textbf{86.89} & 82.79 & 85.55 & 78.18 & \textbf{84.50} &
76.99 & \textbf{89.51} & 82.83 & 84.30 \\
Kimi-K2.6 &
84.95 & \textbf{84.09} & \textbf{87.86} & 81.82 & 81.01 &
79.92 & 86.48 & \textbf{83.73} & 84.00 \\
Qwen3-VL-235B-A22B &
65.05 & 63.96 & 60.69 & 70.91 & 63.95 &
63.18 & 69.46 & 57.53 & 64.00 \\
Qwen3-VL-8B &
51.94 & 43.18 & 36.99 & 58.18 & 39.53 &
38.08 & 49.42 & 40.66 & 43.80 \\
InternVL3.5-241B-A28B &
28.64 & 6.82 & 17.34 & 27.27 & 18.22 &
15.90 & 20.28 & 14.16 & 17.20 \\
InternVL3.5-8B &
8.25 & 2.60 & 6.36 & 12.73 & 7.75 &
6.28 & 7.23 & 5.12 & 6.30 \\
\bottomrule
\end{tabular}}
\caption{Accuracy (\%) on \benchmark{} by problem type and reasoning depth.
DC: DC Networks; AC: AC \& Phasor; T/F: Transient \& Frequency; Sym.:
Symbolic \& Signal; Mix: Mixed \& General; Multi: Multi-Step; Long:
Long-Horizon. Overall averages all 1,000 problems. Bold marks each column's
highest fixed-benchmark point estimate, not statistically significant
superiority.}
\label{tab:main-results}
\end{table*}

\subsection{Main Results}

Table~\ref{tab:main-results} yields three main findings.
\textbf{First, the benchmark remains far from saturated.} At the point-estimate
level, GPT-5.6-sol obtains 84.80\%, followed by Qwen3.5-122B-A10B at 84.30\%
and Kimi-K2.6 at 84.00\%; these small fixed-benchmark gaps are descriptive, not
claims of statistical distinguishability. The highest-scoring system still
misses 152 problems, and the leading open-source MLLMs are competitive with
the strongest commercial chatbot systems.

\textbf{Second, group comparisons depend on model selection.} The three
commercial chatbot systems average 77.17\%, versus 77.43\% for the three
strongest open-source MLLMs and 49.93\% for all six, including compact,
low-scoring baselines. A single group mean therefore conceals both the
competitive open-source frontier and its internal spread.

\textbf{Third, long-horizon dependency remains a stable bottleneck.} Every
model scores below its Multi-Step result on Long-Horizon: drops for GPT-5.6-sol,
Gemini 3.1 Pro, Claude Sonnet 4.6, Qwen3.5, Kimi-K2.6, and Qwen3-VL-235B are
7.99, 7.23, 10.42, 6.68, 2.75, and 11.93 points. Because generation length
remains available, a short output ceiling does not explain this pattern.

\subsection{Performance by Problem Type}

The type breakdown reveals complementary strengths. GPT-5.6-sol leads Symbolic
\& Signal (90.91\%), Kimi-K2.6 leads AC \& Phasor (84.09\%) and Transient \&
Frequency (87.86\%), and Qwen3.5 leads Mixed \& General (84.50\%).
Qwen3-VL-8B falls to 36.99\% on Transient \& Frequency and 39.53\% on Mixed \&
General; both InternVL configurations are weakest on AC \& Phasor. These
regimes stress complex quantities, references, peak/RMS distinctions, and
signs, while Figure~\ref{fig:diagnostic-radar} exposes additional profile
variation in topology, equations, propagation, and physical conventions.

\subsection{Long-Horizon Failure Modes}

Qualitative inspection reveals three recurring error patterns:
\textbf{topology-to-target binding} (wrong quantity; \texttt{000298}:
\(1\,\mathrm{A}\) intermediate versus the requested \(3\,\mathrm{A}\) downward
in the shared \(2\,\Omega\) branch); \textbf{convention propagation} (lost
convention; \texttt{000691}: no one-half gives \(25\,\mathrm{W}\) versus
\(12.5\,\mathrm{W}\)); and \textbf{late-stage completion} (unfinished chain;
\texttt{000174}: three outputs, final loading missing).
Complete diagrams, representative model responses, Gold solutions, and
annotated error traces are provided in the appendix.

\section{Evaluation Reliability}
\label{sec:reliability}

We validate both rule and transport layers. In a 420-decision audit, the
deterministic scorer and two independent LLM checks agreed on 398 decisions;
direct rechecking confirmed the deterministic result in all 22 disagreements.
The scorer also passes 96 regression cases spanning tolerances, units,
multi-slot completeness, signs, phases, and references.

Replaying JSON/LaTeX recovery across 69 repairs in 11 runs produced zero
deterministic-verdict or final-consensus changes; repair applies only when
strict JSON parsing succeeds. Identity blinding, independent votes, strict
majority, and requested-part checks further reduce dependence on one judge.

\section{Limitations}

\benchmark{} focuses on university-level analysis and final-answer accuracy; it
excludes broader electronics and power systems, interactive editing, and
intermediate-state scoring. Worked solutions support future process metrics.

\section{Conclusion}

\benchmark{} pairs 1,000 authentic problems with typed answers and worked
solutions. Across nine models, accuracy peaks at 84.8\%, yet all decline from
Multi-Step to Long-Horizon, exposing a gap between component recognition and
coherent topology-to-equation reasoning.

\bibliography{circuitreason}

\end{document}